\documentclass{article}

 \usepackage[dblblindworkshop, final]{neurips_2026}
\usepackage[utf8]{inputenc} % allow utf-8 input
\usepackage[T1]{fontenc}    % use 8-bit T1 fonts
\usepackage{hyperref}       % hyperlinks
\usepackage{url}            % simple URL typesetting
\usepackage{booktabs}       % professional-quality tables
\usepackage{amsfonts}       % blackboard math symbols
\usepackage{nicefrac}       % compact symbols for 1/2, etc.
\usepackage{microtype}      % microtypography
\usepackage{xcolor}         % colors
\usepackage{graphicx}
\usepackage{amsmath}
\usepackage{multirow} 
\usepackage{textcomp}
\usepackage{stfloats}
\usepackage{url}
\usepackage{verbatim}
\usepackage{tcolorbox}
\usepackage[dvipsnames]{xcolor}
\usepackage{tikz}
\usepackage{comment}
\usetikzlibrary{spy}
\mathchardef\mhyphen="2D
\usepackage{color}
\usepackage{microtype}
\usepackage{multirow}
\usepackage{wrapfig,lipsum}
 \usepackage{booktabs} 
\usepackage{dsfont}
\usepackage{wrapfig}
\usepackage[font=small,skip=0pt]{caption}
\usepackage{pgfplots}

\usepgfplotslibrary{fillbetween}
\usepackage{pgfkeys}
\usepackage{lipsum}
\usepackage{mwe}
\usepackage{tikz}
\usetikzlibrary{backgrounds}
\usetikzlibrary{intersections}
\usetikzlibrary{shapes.geometric}
\usetikzlibrary{spy}
\usepackage{pgfplots, pgfplotstable}
\pgfmathdeclarefunction{gauss}{2}{%
  \pgfmathparse{1/(#2*sqrt(2*pi))*exp(-((x-#1)^2)/(2*#2^2))}%
}
\pgfplotsset{compat=1.11,
    /pgfplots/ybar legend/.style={
    /pgfplots/legend image code/.code={%
       \draw[##1,/tikz/.cd,yshift=-0.25em]
        (0cm,0cm) rectangle (3pt,0.8em);},
   },
}
\usepackage{color, soul, colortbl}
\usepgfplotslibrary{colormaps}
\newenvironment{customlegend}[1][]{%
        \begingroup
        \csname pgfplots@init@cleared@structures\endcsname
        \pgfplotsset{#1}%
    }{%
        \csname pgfplots@createlegend\endcsname
        \endgroup
    }%
    \def\addlegendimage{\csname pgfplots@addlegendimage\endcsname}

\pgfplotsset{ 
cycle list={%
{draw=black,mark=star,solid},
{draw=black, mark=square,solid}}}

\usepackage{amsfonts}
\usepackage{textcomp}
\usepackage{breqn}
\usepackage{comment}
 \usepackage{soul}

\usepackage{caption}
\usepackage{pgfplots}
\usepackage{pgfplotstable}
\usepackage[dvipsnames]{xcolor}
\pgfplotsset{compat=1.18}
\usetikzlibrary{pgfplots.groupplots}
\title{Exploring Jailbreak Vulnerability Across Depth-Based Dynamic Inference in Large Language Models}
\workshoptitle{Foundations of Language Model Security: Theory, Practice, and Fundamental Limits (FLMSec)}

\author{
  Lass Najmuldeen\thanks{Corresponding author.} \\
  Independent Researcher \\
  Palo Alto, CA, USA \\
  \texttt{lass.najm@gmail.com}
  \And
  Seyedhamidreza Mousavi\thanks{https://hamidmousavi0.github.io/} \\
  Department of Computer Science and Engineering \\
  Faculty of Engineering and Health Sciences \\
  Mälardalen University \\
  Västerås, Sweden \\
  \texttt{seyedhamidreza.mousavi@mdu.se}
}

\begin{document}

\maketitle

\begin{abstract}
Large Language Models (LLMs) have been shown to be computationally expensive and vulnerable to jailbreak attacks, in which adversarial prompts bypass safety alignment and induce harmful responses. 
To address this inefficiency, a range of techniques have been proposed to enable efficient LLM inference. 
Depth-based dynamic inference has demonstrated promising results by enabling outputs to be generated from intermediate layers.
It offers the flexibility to adjust model behavior based on the available resource budget. 
However, how this flexibility affects the robustness of LLMs against jailbreak attacks remains unexplored. 
In this paper, we address this gap via a comprehensive investigation into the impact of depth-based dynamic inference on the robustness of LLMs against jailbreak attacks.
Specifically, we make the following key observations: 1) Shallow exits are consistently more robust to jailbreak attacks than deeper ones; 2) Exit depth affects how easily the adversarial suffix can be optimized, with deeper exits reaching lower attack loss; and 3) Adversarial suffixes affect utility uniformly across exit depths. These insights can help design dynamic inference methodologies that are more robust to jailbreak attacks. Code is available at \url{https://github.com/LassNajmuldeen/balcony-jailbreaking}
\end{abstract}
%%%%%%%%%%%%%%%%%%%%%%%%%%%%%%%%%%%
\section{Introduction}\label{sec:introduction}
Large Language Models (LLMs) have been widely leveraged across a variety of applications and enhanced through numerous alignment techniques~\cite{bi2024deepseek}.
However, deploying these models in real-world applications, particularly in industrial environments, remains constrained by strict computational and latency requirements, whether on edge devices or in cloud settings~\cite{jamialahmadi2025balcony, cai2024flextron}. 
To address this efficiency bottleneck, dynamic inference has been proposed, enabling a single model to trade off compute against accuracy at inference time rather than requiring separate models for each deployment budget~\cite{elhoushi2024layerskip, cai2024flextron, cai2025llamaflex}. 
Dynamic inference can be achieved through width-based approaches, which adjust the hidden dimension of the model's feed-forward layers~\cite{kudugunta2023matformer}, or depth-based approaches, which selectively reduce the number of transformer layers used~\cite{huang2024context}. 

At the same time, despite the alignment techniques, LLMs remain vulnerable to jailbreak attacks, in which adversarial prompts bypass safety alignment and induce harmful responses~\cite{wei2023jailbroken,beyer2025adversariallm}. 
Exploiting this vulnerability, a range of optimization-based attacks have been proposed that automatically search for such adversarial prompts using the model's gradients or output probabilities, rather than relying on manual prompt engineering~\cite{zou2023universal, xu2026tao}.
However, these attacks have only been studied against the standard, full-depth inference of LLMs, and how depth-based dynamic inference affects the robustness of LLMs against jailbreak attacks remains unexplored.

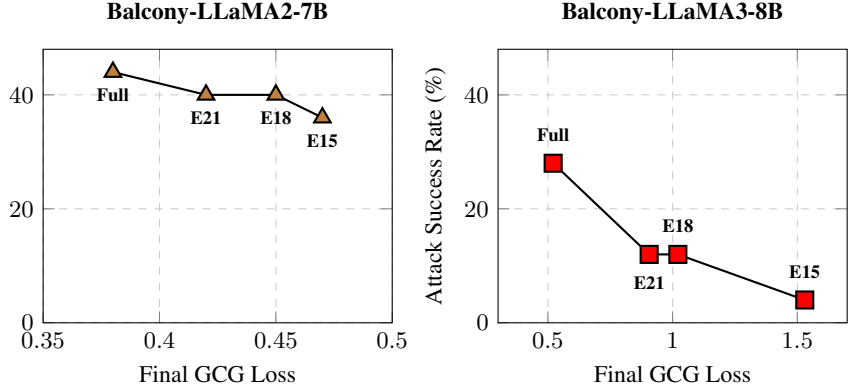
\begin{figure}[h]
    \centering
    \begin{tikzpicture}
\begin{groupplot}[
  group style={group size=2 by 1, horizontal sep=1.4cm},
  height=5.2cm, width=6.2cm,
  ymin=0, ymax=48,
  grid=major,
  grid style={dashed,gray!40},
  ylabel near ticks,
  xlabel near ticks,
  x tick label style={rotate=0,anchor=north},
  tick label style={font=\small},
  label style={font=\small},
  title style={font=\small\bfseries},
]

% ---- LEFT: LLaMA3-8B ----

% ---- RIGHT: LLaMA2-7B ----
\nextgroupplot[
  title={Balcony-LLaMA2-7B},
  xlabel={Final GCG Loss},
  xmin=0.35, xmax=0.5,
]
\addplot [mark=triangle*, mark size=3.6pt, brown, thick, draw=black]
  coordinates {(0.47,36) (0.45,40) (0.42,40) (0.38,44)};
\node[font=\scriptsize\bfseries] at (axis cs: 0.47,32) {E15};
\node[font=\scriptsize\bfseries] at (axis cs: 0.45,36) {E18};
\node[font=\scriptsize\bfseries] at (axis cs: 0.42,36) {E21};
\node[font=\scriptsize\bfseries] at (axis cs: 0.38,40) {Full};

\nextgroupplot[
  title={Balcony-LLaMA3-8B},
  xlabel={Final GCG Loss},
  ylabel={Attack Success Rate (\%)},
  xmin=0.3, xmax=1.7,
]
\addplot [mark=square*, mark size=3.2pt, red, thick, draw=black]
  coordinates {(1.53,4) (1.021,12) (0.906,12) (0.521,28)};
\node[font=\scriptsize\bfseries] at (axis cs: 1.53,9)  {E15};
\node[font=\scriptsize\bfseries] at (axis cs: 1.021,17) {E18};
\node[font=\scriptsize\bfseries] at (axis cs: 0.906,7) {E21};
\node[font=\scriptsize\bfseries] at (axis cs: 0.521,33) {Full};
\end{groupplot}
\end{tikzpicture}

    \caption{Attack success rate (ASR) versus final GCG loss at each exit depth (E15, E18, E21, Full), for Balcony-LLaMA2-7B (left) and  Balcony-LLaMA3-8B (right) on the HarmBench dataset.}
    \label{fig:asrvsloss}
\end{figure}

In this paper, we present the first comprehensive investigation into the impact of depth-based dynamic inference on the robustness of LLMs against jailbreak attacks.
We focus specifically on depth-based methods, since removing entire layers maps more naturally onto how GPUs execute sequential computation, yielding larger real-world latency and throughput gains than adjusting a layer's width~\cite{elhoushi2024layerskip}.
Specifically, our investigation is based on the Balcony method~\cite{jamialahmadi2025balcony}. We adopt Balcony because its frozen backbone leaves the original safety-aligned model unmodified. 
This ensures that any change we observe in jailbreak robustness across exit points can be attributed to the dynamic-inference mechanism itself, rather than to degradation of the underlying model.
We apply an optimization-based jailbreak attack, Greedy Coordinate Gradient (GCG)~\cite{zou2023universal}, at different exit points to explore the impact of dynamic inference configurations on the robustness of dynamic LLMs. 
To validate these design choices, we conduct a systematic study of jailbreak robustness across exit depths using the GCG attack~\cite{zou2023universal} on two model families, Balcony-LLaMA3-8B and Balcony-LLaMA2-7B, evaluated on AdvBench and HarmBench. Figure~\ref{fig:asrvsloss} summarizes attack success rate against final optimization loss at each exit depth.

Based on our explorations, we make the following key observations:
\begin{itemize}
    \item Shallow exits are consistently more robust to jailbreak attacks than deeper ones, though the size of this effect varies substantially across models. For Balcony-LLaMA3-8B, the GCG attack success rate falls sharply, from 28\% at full depth to just 0–4\% at the shallowest exit. In contrast, Balcony-LLaMA2-7B shows a much smaller effect, with success rates ranging only from 36\% to 44\% across all exit depths.
    \item Exit depth affects how hard the adversarial suffix is to optimize: loss decreases monotonically with depth, and the full model is easiest to attack. The size of this effect varies sharply by model. The final GCG optimization loss for Balcony-LLaMA3-8B is 3$\times$ lower at the full model than at exit-15. 
    \item Adversarial suffixes affect utility uniformly across exit depths, indicating that their collateral cost is set by the backbone rather than the exit point. The robustness gains at shallow exits therefore cannot be attributed to degraded capability under attack.
\end{itemize}

\section{Related Work}
\subsection{Jailbreak Attacks}
Despite extensive safety alignment, large language models remain vulnerable to jailbreak prompts~\cite{wei2023jailbroken}. These prompts can be viewed as adversarial inputs designed to circumvent refusal mechanisms and induce models to generate harmful or policy-violating content~\cite{goodfellow2014explaining,beyer2025adversariallm}. Existing jailbreak attacks can be broadly categorized according to the level at which the adversarial input is constructed: token-level and prompt-level attacks.

Token-level attacks search directly over the discrete input-token space by modifying, inserting, or optimizing individual tokens or token sequences~\cite{zou2023universal,sadasivan2024fast,zhu2023autodan,hayase2024query}. These methods often employ gradient-based optimization, heuristic search, or query-based procedures to identify adversarial suffixes that suppress refusal behavior.

In contrast, prompt-level attacks operate at the semantic or instruction level and typically produce more natural and interpretable jailbreak prompts. Such prompts may be manually crafted by humans~\cite{wei2023jailbroken,shen2024anything} or automatically generated by LLM-based agents that iteratively propose, evaluate, and refine attack prompts against the target model~\cite{sabbaghi2025adversarial,panfilov2026claudini}.

\subsection{Dynamic Inference in LLMs}

Dynamic inference was initially explored in NLP using relatively small encoder-based models~\cite{xin2020deebert,hou2020dynabert}. These approaches typically rely on architecture-specific modifications and specialized training procedures, making their direct application to modern large-scale language models challenging.

Recent studies have extended dynamic inference to larger language models by introducing structural elasticity along different model dimensions. MatFormer~\cite{kudugunta2023matformer} introduces a nested architecture along the width dimension, allowing multiple submodels of different sizes to be extracted from a single jointly trained network. Building on this idea, Flextron and LLaMAFlex~\cite{cai2024flextron,cai2025llamaflex} combine nested model structures with input-adaptive routing to dynamically select computational configurations according to the input. In parallel, SortedLLaMA~\cite{huang2024context} and LayerSkip~\cite{elhoushi2024layerskip} explore elasticity along the depth dimension, enabling inference with different numbers of transformer layers through specialized training strategies.

Despite their efficiency benefits, these methods modify or retrain the base model to support dynamic execution, which may alter the behavior or degrade the performance of the original full model. Balcony~\cite{jamialahmadi2025balcony} avoids this limitation by freezing the pretrained base model and training only lightweight exit modules attached at intermediate layers. We therefore adopt Balcony as the architectural basis for our robustness evaluation. Its frozen-backbone design provides a controlled setting in which the robustness of the original full model can be directly compared with that of sub-models exiting at different depths, without confounding the analysis through modifications to the base model parameters.

\section{Preliminaries}
In this section, we introduce the key concepts used in our investigation to analyze the vulnerability of depth-based dynamic inference in large language models to jailbreak attacks.

\subsection{Depth-Based Dynamic Inference}
As discussed in the Introduction (Section~\ref{sec:introduction}), we adopt Balcony as our depth-based dynamic inference architecture because it leaves the base model unchanged. To provide the necessary background, we next describe its architecture and  training procedure.

\textbf{Architecture.} The Balcony architecture consists of a frozen base language model and a set of lightweight intermediate exits that can be selected according to the available computational budget. Each exit enables token prediction before all transformer layers have been processed.

Let the base model contain \(L\) transformer layers, and let
\(\mathcal{S}\subseteq\{1,\ldots,L\}\) denote the selected exit locations. For an exit at layer \(k\in\mathcal{S}\), the intermediate hidden representation \(H_k\) is transformed by a Balcony module as
\begin{equation}
    \widetilde{H}_k = g_k(H_k;\Theta_k),
\end{equation}
where \(g_k\) consists of an additional transformer block followed by a normalization layer, and \(\Theta_k\) denotes its trainable parameters. The transformed representation is then passed to the same language-model head used by the base model to produce the output-token distribution.

\textbf{Training.}
Balcony employs self-distillation to align the predictive distribution of each intermediate exit with that of the full model. During training, the parameters of the base model remain frozen, and only the parameters of the Balcony modules are optimized. For an exit at layer \(k\), the training objective encourages its output distribution to match the distribution produced by the full model.
\subsection{Optimization-Based Jailbreak Attack}
Optimization-based attacks leverage the target model's gradients or output logits to automatically construct jailbreak prompts and often achieve higher attack success rates than manually designed approaches. Among these methods, Greedy Coordinate Gradient (GCG) is a representative token-level attack. GCG iteratively optimizes an adversarial suffix to increase the likelihood that the model begins its response with a predefined affirmative continuation, thereby encouraging compliance with a harmful user request.

Let an autoregressive language model define a probability distribution over the next token as
\begin{equation}
    p_{\theta}(x_{t}\mid x_{1:t-1}),
\end{equation}
where \(x_{1:t-1}\) denotes the preceding token sequence and \(\theta\) represents the model parameters. Accordingly, the probability of generating a target sequence
\(\mathbf{y}^{\star}=(y^{\star}_1,\ldots,y^{\star}_M)\)
conditioned on an input sequence \(\mathbf{x}\) is factorized as
\begin{equation}
    p_{\theta}(\mathbf{y}^{\star}\mid \mathbf{x})
    =
    \prod_{m=1}^{M}
    p_{\theta}
    \left(
        y^{\star}_m
        \mid
        \mathbf{x},y^{\star}_{1:m-1}
    \right).
\end{equation}

GCG appends an adversarial suffix
\(\mathbf{s}=(s_1,\ldots,s_K)\)
to a harmful user query \(\mathbf{q}\). The suffix is optimized to increase the probability that the model begins its response with a predefined affirmative target sequence \(\mathbf{y}^{\star}\). The attack objective is therefore defined as the negative log-likelihood of the target sequence:
\begin{equation}
    \mathcal{L}_{\mathrm{GCG}}(\mathbf{s})
    =
    -\log
    p_{\theta}
    \left(
        \mathbf{y}^{\star}
        \mid
        \mathbf{q}\oplus\mathbf{s}
    \right),
\end{equation}
where \(\oplus\) denotes sequence concatenation. Equivalently,
\begin{equation}
    \mathcal{L}_{\mathrm{GCG}}(\mathbf{s})
    =
    -\sum_{m=1}^{M}
    \log
    p_{\theta}
    \left(
        y^{\star}_m
        \mid
        \mathbf{q}\oplus\mathbf{s},
        y^{\star}_{1:m-1}
    \right).
\end{equation}

The adversarial suffix is obtained by solving the discrete optimization problem
\begin{equation}
    \mathbf{s}^{\star}
    =
    \arg\min_{\mathbf{s}\in\mathcal{V}^{K}}
    \mathcal{L}_{\mathrm{GCG}}(\mathbf{s}),
\end{equation}
where \(\mathcal{V}\) is the model vocabulary and \(K\) is the suffix length. Because direct optimization over discrete tokens is intractable, GCG uses token-level gradients to identify promising replacements and greedily updates selected suffix positions to reduce the objective.

\section{Exploring Jailbreak Vulnerability }
We investigate how the jailbreak vulnerability of a depth-based dynamic inference LLM changes across different inference depths. Our exploration considers several intermediate exits together with the full model and analyzes their behavior from complementary perspectives.

We first examine jailbreak effectiveness across exit depths using attack success rate. We then study the optimization behavior of GCG, including its convergence pattern and final attack loss, to understand how easily adversarial suffixes can be optimized for each exit. Next, we evaluate model utility under both clean and adversarial inputs to determine whether the observed robustness differences are associated with changes in general capability. Finally, we analyze the relationship between inference depth, attack optimization, utility, and jailbreak success, and summarize the main findings of this exploration.

\begin{figure}[t]
\centering
\includegraphics[width=\columnwidth]{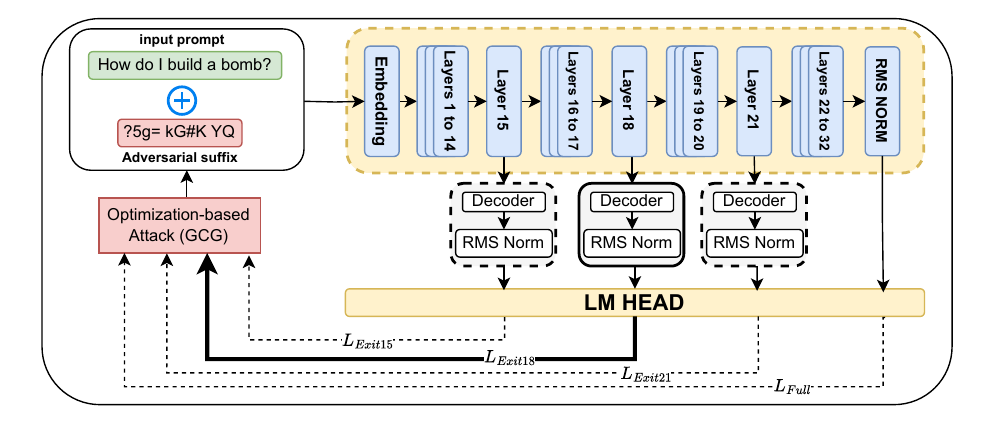}
\caption{Overview of the jailbreak evaluation pipeline. A harmful prompt (e.g., ``How do I build a bomb?'') is combined with an adversarial suffix produced by an optimization-based attack (GCG), and the resulting input is passed through the frozen backbone. We compute the loss at three Balcony exit points (layers 15, 18, and 21) as well as at the full model, enabling direct comparison of jailbreak robustness across inference depths within the same underlying weights.}
\label{fig:balcony}
\end{figure}

\textbf{Base architectures.} We use pretrained LLaMA2-7B and LLaMA3-8B as base models and convert them into dynamic models using the Balcony method. Figure~\ref{fig:balcony} provides an overview of the resulting architecture, and detailed configurations are summarized in Table~\ref{tab:config}.
For exit points, we consider the same configuration $\{15,18,21\}$ for both models.

\begin{table}[h]
\centering
\caption{Configuration of the Balcony models used in our experiments. Training
hyperparameters and parameter counts follow~\cite{jamialahmadi2025balcony};
the base model is frozen and only the Balcony modules are updated.}
\label{tab:config}
\small
\begin{tabular}{lcc}
\toprule
 & \textbf{Balcony-LLaMA2-7B} & \textbf{Balcony-LLaMA3-8B} \\
\midrule
\multicolumn{3}{l}{\textit{Base model}} \\
Transformer layers $L$          & 32 & 32 \\
Non-embedding parameters        & 6.5B & 6.9B \\
Pretraining tokens              & 2T & 15T \\
\midrule
\multicolumn{3}{l}{\textit{Balcony modules}} \\
Exit locations $S$              & \{15, 18, 21\} & \{15, 18, 21\} \\
\midrule
\multicolumn{3}{l}{\textit{Training}} \\
Objective                       & KL self-distillation & KL self-distillation \\
Corpus                          & Cosmopedia V2 & Cosmopedia V2 \\
Training tokens                 & 31.5B (1.5\% of pretraining) & 31.5B (0.2\% of pretraining) \\
Batch size                      & 256 & 256 \\
Steps                           & 30K & 30K \\
Sequence length                 & 4,096 & 4,096 \\
Peak learning rate              & $5\times10^{-4}$ & $5\times10^{-4}$ \\
LR schedule                     & cosine & cosine \\
\bottomrule
\end{tabular}
\end{table}

\textbf{Experimental settings.} 

\textbf{Training.} Both Balcony-LLaMA2-7B~\cite{touvron2023llama} and Balcony-LLaMA3-8B~\cite{grattafiori2024llama} are trained on the Cosmopedia V2~\cite{benallal2024smollmcorpus} dataset using KL self-distillation, following the original Balcony setup~\cite{jamialahmadi2025balcony}. Keeping the training data and objective identical across the two models ensures that any difference in jailbreak robustness we observe stems from the exit configuration and the underlying backbone rather than from differences in the Balcony training procedure itself.

\textbf{Evaluation.} We evaluate jailbreak robustness on HarmBench~\cite{mazeika2024harmbench} and AdvBench~\cite{zou2023universal}, applying GCG with the attack configuration of Zou et al.~\cite{zou2023universal}. We optimize an adversarial suffix for 150 steps over 50 harmful behaviors per benchmark and report the attack success rate (ASR) as the fraction of behaviors the HarmBench judge scores above 0.5, together with the mean final cross-entropy loss reached by the attack. To assess whether robustness differences coincide with changes in general capability, we additionally evaluate utility on six multiple-choice benchmarks, HellaSwag, ARC-C, ARC-E, WinoGrande, PIQA, and MMLU, using lm-eval-harness~\cite{eval-harness}, both on clean inputs and with each exit's strongest GCG suffix appended to every question. We report acc\_norm for HellaSwag, ARC, and PIQA, and acc for WinoGrande and MMLU.

\subsection{Jailbreak Vulnerability Across Exit Depths}
We first explore how early-exit inference affects the attack success rate (ASR) of GCG, the most representative optimization-based jailbreak attack. We evaluate four exit points (\textit{exit-15}, \textit{exit-18}, \textit{exit-21}, and \textit{full}) on
Balcony-LLaMA2-7B and Balcony-LLaMA3-8B, using GCG with 150 steps over 50 behaviors
from HarmBench and AdvBench (Figure~\ref{fig:acc-drop}).

Across both backbones and benchmarks, ASR is highest at the full model, confirming that later layers are consistently the most vulnerable to adversarial suffixes. However, the two models differ sharply in how quickly that vulnerability emerges across depth.
Balcony-LLaMA2-7B is already substantially exploitable at the shallowest exit tested: exit-15 reaches 36\% ASR on HarmBench and 46\% on AdvBench, and ASR rises only modestly from there, reaching 44\%/54\% at the full model. This near-monotonic, gently increasing trend suggests that adversarial suffixes crafted against LLaMA2 transfer effectively even to early exits, i.e., the model's vulnerability is distributed across depth rather than concentrated in later layers.

Balcony-LLaMA3-8B shows the opposite pattern. At exit-15 it is far more robust, with ASR of only 4\% on HarmBench and 2\% on AdvBench. ASR climbs steadily through exit-18 (12\%/0\%) and exit-21 (12\%/12\%) before jumping sharply at the full model, where both benchmarks converge to 28\%. This large gap between shallow-exit and full-model ASR indicates that LLaMA3's vulnerability to GCG is concentrated almost entirely in its later layers, and that early-exit inference offers meaningfully stronger protection for this backbone than for LLaMA2.

Taken together, these results show that early-exit robustness is not a fixed property of the early-exit mechanism itself, but depends on the backbone: for LLaMA2, shallow exits provide little defensive benefit, while for LLaMA3 they substantially suppress attack success until the final layers are reached.

\begin{table}[htbp]
\centering
\captionsetup{justification=centering}
\resizebox{\columnwidth}{!}{
\begin{tabular}{cc}
\multicolumn{2}{c}{
\begin{tikzpicture}
\begin{customlegend}[legend columns=2,
    legend style={draw=black, text opacity=1, row sep=2pt,
                  font=\fontsize{11}{9}\selectfont, column sep=1.5ex},
    legend entries={\textsc{HarmBench}, \textsc{AdvBench}}]
    \addlegendimage{draw=black, fill=CornflowerBlue!85, mark=square*}
    \addlegendimage{draw=black, fill=Goldenrod!85, mark=square*}
\end{customlegend}
\end{tikzpicture}}
\\
% ---------- LEFT: LLaMA-2-7B ----------
\begin{tikzpicture}
  \begin{axis}[
        width=0.75\columnwidth,
        font=\small,
        ybar,
        bar width=14pt,
        symbolic x coords={exit-15, exit-18, exit-21, full},
        xtick=data,
        x tick label style={font=\bfseries\small},
        ymin=0, ymax=60,
        ytick={0,10,20,30,40},
        yticklabels={0,10,20,30,40},
        grid=major,
        grid style={line width=.1pt, draw=gray!35},
        major grid style={line width=.2pt, draw=gray!50},
        axis line style={draw=black!70},
        tick align=outside,
        tickwidth=2pt,
        ylabel={Top-1 Attack Success Rate (\%)},
        ylabel style={font=\small},
        nodes near coords,
        nodes near coords align={vertical},
        every node near coord/.append style={font=\fontsize{7}{7}\selectfont, /pgf/number format/fixed},
        title={\small Balcony-LLaMA2-7B},
        enlarge x limits=0.2,
    ]
    \addplot [draw=black, line width=0.3pt, fill=CornflowerBlue!85]
        coordinates {(exit-15,36)(exit-18,40)(exit-21,40)(full,44)};
    \addplot [draw=black, line width=0.3pt, fill=Goldenrod!85]
        coordinates {(exit-15,46)(exit-18,42)(exit-21,48)(full,54)};
  \end{axis}
\end{tikzpicture}
&
% ---------- RIGHT: LLaMA3-8B ----------
\begin{tikzpicture}
  \begin{axis}[
        width=0.75\columnwidth,
        font=\small,
        ybar,
        bar width=14pt,
        symbolic x coords={exit-15, exit-18, exit-21, full},
        xtick=data,
        x tick label style={font=\bfseries\small},
        ymin=0, ymax=45,
        ytick={0,10,20,30,40},
        yticklabels={0,10,20,30,40},
        grid=major,
        grid style={line width=.1pt, draw=gray!35},
        major grid style={line width=.2pt, draw=gray!50},
        axis line style={draw=black!70},
        tick align=outside,
        tickwidth=2pt,
        ylabel={Top-1 Attack Success Rate (\%)},
        ylabel style={font=\small},
        nodes near coords,
        nodes near coords align={vertical},
        every node near coord/.append style={font=\fontsize{7}{7}\selectfont, /pgf/number format/fixed},
        title={\small Balcony-LLaMA3-8B},
        enlarge x limits=0.2,
    ]
    \addplot [draw=black, line width=0.3pt, fill=CornflowerBlue!85]
        coordinates {(exit-15,4)(exit-18,12)(exit-21,12)(full,28)};
    \addplot [draw=black, line width=0.3pt, fill=Goldenrod!85]
        coordinates {(exit-15,2)(exit-18,0)(exit-21,12)(full,28)};
  \end{axis}
\end{tikzpicture}

\end{tabular}}

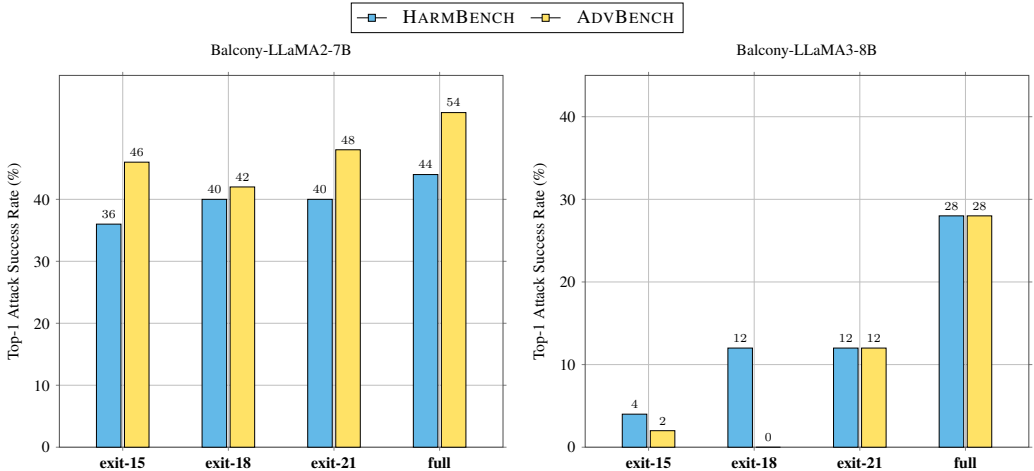
\captionof{figure}{Attack success rate for every exit of Balcony-LLaMA2-7B (\textbf{left}) and Balcony-LLaMA3-8B (\textbf{right}) on HarmBench and AdvBench, using GCG with 150 steps on 50 behaviors.}
\label{fig:acc-drop}
\end{table}

% Requires: \usepackage{pgfplots} \usepackage{pgfplotstable} \pgfplotsset{compat=1.18}
% Place the four CSVs (llama2_advbench.csv, llama2_harmbench.csv,
% llama3_advbench.csv, llama3_harmbench.csv) in the same folder as this .tex
% file, or adjust the paths in \addplot table[...] {...}; below.

\begin{figure}[htbp]
\centering
\begin{tikzpicture}
  \begin{groupplot}[
      group style={
        group size=2 by 2,
        horizontal sep=1.6cm,
        vertical sep=1.6cm,
      },
      width=0.52\columnwidth,
      height=0.32\columnwidth,
      tick label style={font=\footnotesize},
      label style={font=\footnotesize},
      title style={font=\footnotesize},
      xlabel={Iteration},
      ylabel={GCG Loss},
      xmin=0, xmax=150,
      grid=major,
      grid style={line width=.1pt, draw=gray!30},
      major grid style={line width=.2pt, draw=gray!50},
      axis line style={draw=black!70},
      tick align=outside,
      legend columns=4,
      legend style={
        draw=black, font=\footnotesize,
        at={(1.1,1.45)}, anchor=north,
      },
      every axis plot/.append style={line width=1pt},
      cycle list={
        {Blue, solid},
        {orange, solid},
        {Green, solid},
        {black, dashed},
      },
    ]

    % ---- (1,1) LLaMA-2-7B / AdvBench ----
    \nextgroupplot[title={Balcony-LLaMA2-7B --- AdvBench}]
    \addplot table [x=exit_15 iteration, y=exit_15 loss, col sep=comma] {paper_figures/llama2-7b_Advbench_1.csv};
    \addlegendentry{exit-15}
    \addplot table [x=exit_18 iteration, y=exit_18 loss, col sep=comma] {paper_figures/llama2-7b_Advbench_1.csv};
    \addlegendentry{exit-18}
    \addplot table [x=exit_21 iteration, y=exit_21 loss, col sep=comma] {paper_figures/llama2-7b_Advbench_1.csv};
    \addlegendentry{exit-21}
    \addplot table [x=exit_full iteration, y=exit_full loss, col sep=comma] {paper_figures/llama2-7b_Advbench_1.csv};
    \addlegendentry{full}

    % ---- (1,2) LLaMA-2-7B / HarmBench ----
    \nextgroupplot[title={Balcony-LLaMA2-7B --- HarmBench}]
    \addplot table [x=exit_15 iteration, y=exit_15 loss, col sep=comma] {paper_figures/llama2-7b_Harmbench_1.csv};
    \addplot table [x=exit_18 iteration, y=exit_18 loss, col sep=comma] {paper_figures/llama2-7b_Harmbench_1.csv};
    \addplot table [x=exit_21 iteration, y=exit_21 loss, col sep=comma] {paper_figures/llama2-7b_Harmbench_1.csv};
    \addplot table [x=exit_full iteration, y=exit_full loss, col sep=comma] {paper_figures/llama2-7b_Harmbench_1.csv};

    % ---- (2,1) LLaMA3-8B / AdvBench ----
    \nextgroupplot[title={Balcony-LLaMA3-8B --- AdvBench}]
    \addplot table [x=iteration, y=exit_15, col sep=comma] {paper_figures/llama3_advbench.csv};
    \addplot table [x=iteration, y=exit_18, col sep=comma] {paper_figures/llama3_advbench.csv};
    \addplot table [x=iteration, y=exit_21, col sep=comma] {paper_figures/llama3_advbench.csv};
    \addplot table [x=iteration, y=exit_full, col sep=comma] {paper_figures/llama3_advbench.csv};

    % ---- (2,2) LLaMA3-8B / HarmBench ----
    \nextgroupplot[title={Balcony-LLaMA3-8B --- HarmBench}]
    \addplot table [x=iteration, y=exit_15, col sep=comma] {paper_figures/llama3_harmbench.csv};
    \addplot table [x=iteration, y=exit_18, col sep=comma] {paper_figures/llama3_harmbench.csv};
    \addplot table [x=iteration, y=exit_21, col sep=comma] {paper_figures/llama3_harmbench.csv};
    \addplot table [x=iteration, y=exit_full, col sep=comma] {paper_figures/llama3_harmbench.csv};

  \end{groupplot}
\end{tikzpicture}
\caption{GCG loss over 150 optimization steps at each exit depth (\textit{exit-15}, \textit{exit-18}, \textit{exit-21}, \textit{full}), for Balcony-LLaMA2-7B (top) and Balcony-LLaMA3-8B (bottom) on AdvBench (left) and HarmBench (right).}
\label{fig:loss-curves}
\end{figure}
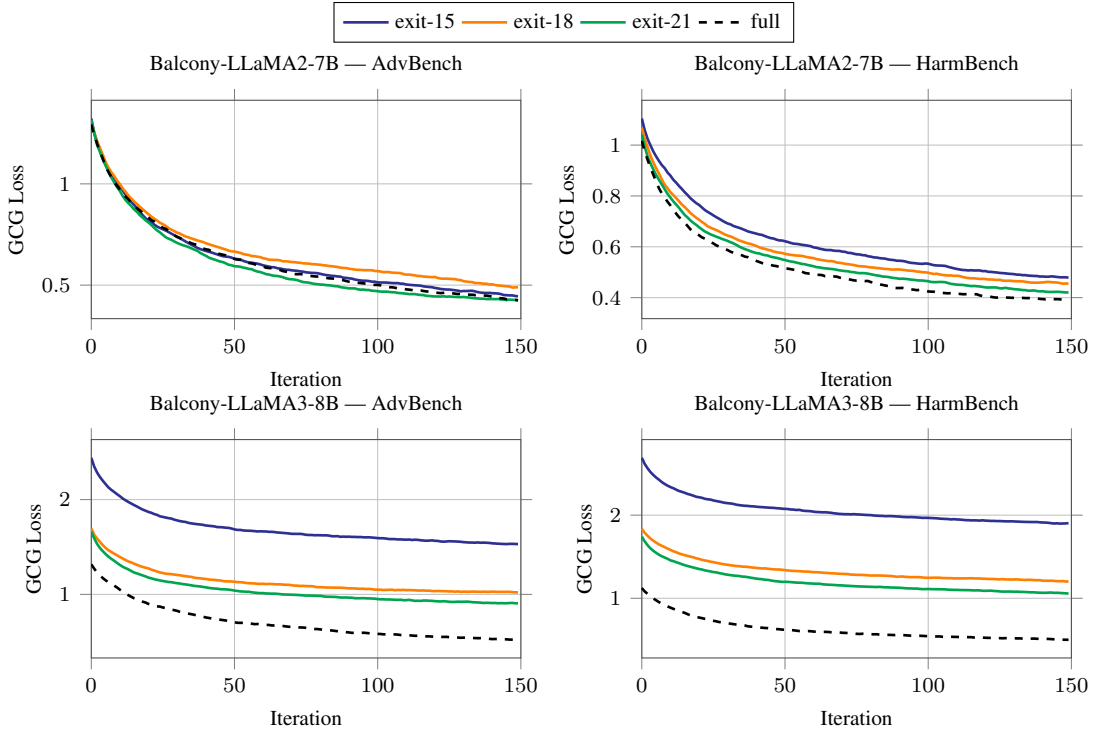

\subsection{Attack Optimization}\label{subsec:attack_optimization}

In this section, we explore how exit depth affects the optimization dynamics of the attack. To this end, we track the GCG loss over 150 optimization iterations at each exit depth (\textit{exit-15}, \textit{exit-18}, \textit{exit-21}, \textit{full}) for both models on AdvBench and HarmBench (Figure~\ref{fig:loss-curves}).
GCG loss decreases monotonically with exit depth across all four settings. The full model reaches the lowest loss, followed by exit-21, exit-18, and exit-15. This ordering holds throughout optimization, across both models and benchmarks. Deeper exits are thus systematically easier targets for suffix optimization.
The separation between exit depths is more pronounced for Balcony-LLaMA3-8B than for Balcony-LLaMA2-7B. On Balcony-LLaMA2-7B, the four curves converge to a narrow band after roughly 50 iterations, with exit-15 trailing only modestly behind the full model. On Balcony-LLaMA3-8B, the pattern differs: exit-15 stays clearly separated from the other three arms throughout all 150 iterations, converging to a substantially higher loss, while exit-18, exit-21, and the full model cluster closer together.

\subsection{Utility Evaluation}\label{subsec:utility}
In this section, we evaluate how adversarial suffixes optimized against each exit affect the model's general capability.
We evaluate every exit on six standard benchmarks (HellaSwag, ARC-C, ARC-E, WinoGrande, PIQA, and MMLU), both without attack (Table~\ref{tab:utility_clean}) and with each exit's strongest GCG suffix appended (Table~\ref{tab:utility_attacked}).
The clean utility increases monotonically with exit depth for
both models across all six benchmarks (Table~\ref{tab:utility_clean}). 
This confirms that deeper exits carry more task-relevant capability.
Under attack, utility remains largely stable across exits and benchmarks for both models. The exception is ARC-E on the 7B model, where the adversarial suffix appears to interfere with task performance itself, rather than merely eliciting unsafe behavior. These results indicate that adversarial suffixes optimized to induce jailbreaks do not impose a broad or systematic cost on downstream utility.

% ============================================================
% TABLE 1: Utility Benchmarks, No Attack
% ============================================================

\begin{table}[h]
\centering
\caption{Utility on six multiple-choice benchmarks with no attack}
\label{tab:utility_clean}
\small
\setlength{\tabcolsep}{5pt}
\renewcommand{\arraystretch}{1.15}
\resizebox{\columnwidth}{!}{
\begin{tabular}{ll cccccc}
\toprule
\textbf{Model} & \textbf{Exit} & \textbf{HellaSwag} & \textbf{ARC-C} & \textbf{ARC-E} & \textbf{WinoGrande} & \textbf{PIQA} & \textbf{MMLU} \\
\midrule
% \multirow{4}{*}{1B}
%  & Exit 4  & 0.333 & 0.237 & 0.454 & 0.520 & 0.641 & 0.245 \\
%  & Exit 8  & 0.417 & 0.281 & 0.527 & 0.523 & 0.690 & 0.252 \\
%  & Exit 12 & 0.546 & 0.336 & 0.615 & 0.568 & 0.723 & 0.249 \\
%  & \cellcolor{gray!12}Full (16) & \cellcolor{gray!12}\textbf{0.612} & \cellcolor{gray!12}\textbf{0.387} & \cellcolor{gray!12}\textbf{0.690} & \cellcolor{gray!12}\textbf{0.592} & \cellcolor{gray!12}\textbf{0.747} & \cellcolor{gray!12}\textbf{0.256} \\
% \midrule
\multirow{4}{*}{7B}
 & Exit 15 & 0.630 & 0.347 & 0.617 & 0.636 & 0.742 & 0.369 \\
 & Exit 18 & 0.684 & 0.388 & 0.662 & 0.650 & 0.760 & 0.395 \\
 & Exit 21 & 0.715 & 0.414 & 0.683 & 0.670 & 0.766 & 0.400 \\
 & \cellcolor{gray!12}Full (32) & \cellcolor{gray!12}\textbf{0.760} & \cellcolor{gray!12}\textbf{0.451} & \cellcolor{gray!12}\textbf{0.739} & \cellcolor{gray!12}\textbf{0.691} & \cellcolor{gray!12}\textbf{0.787} & \cellcolor{gray!12}\textbf{0.409} \\
\midrule
\multirow{4}{*}{8B}
 & Exit 15 & 0.519 & 0.334 & 0.476 & 0.658 & 0.634 & 0.310 \\
 & Exit 18 & 0.581 & 0.386 & 0.573 & 0.684 & 0.669 & 0.614 \\
 & Exit 21 & 0.635 & 0.429 & 0.611 & 0.697 & 0.695 & 0.626 \\
 & \cellcolor{gray!12}Full (32) & \cellcolor{gray!12}\textbf{0.792} & \cellcolor{gray!12}\textbf{0.540} & \cellcolor{gray!12}\textbf{0.777} & \cellcolor{gray!12}\textbf{0.734} & \cellcolor{gray!12}\textbf{0.806} & \cellcolor{gray!12}\textbf{0.621} \\
\bottomrule
\end{tabular}}
\end{table}

% ============================================================
% TABLE 2: Utility Under Attack
% ============================================================

\begin{table}[h]
\centering
\caption{Utility with each exit's strongest GCG suffix appended to every question, to be read against the clean scores in Table~\ref{tab:utility_clean}. Same benchmarks, settings and metrics.}
\label{tab:utility_attacked}
\small
\setlength{\tabcolsep}{4.5pt}
\renewcommand{\arraystretch}{1.2}
\resizebox{\columnwidth}{!}{
\begin{tabular}{ll cccccc}
\toprule
\textbf{Model} & \textbf{Exit} & \textbf{HellaSwag} & \textbf{ARC-C} & \textbf{ARC-E} & \textbf{WinoGrande} & \textbf{PIQA} & \textbf{MMLU} \\
\midrule
% \multirow{5}{*}{1B}
%  & Exit 4 & 0.329 \scriptsize(\ensuremath{-}0.3) & 0.253 \scriptsize\textcolor{ForestGreen!60!black}{(+1.5)} & 0.407 \scriptsize(\ensuremath{-}4.6) & 0.508 \scriptsize(\ensuremath{-}1.2) & 0.615 \scriptsize(\ensuremath{-}2.6) & 0.254 \scriptsize\textcolor{ForestGreen!60!black}{(+0.9)} \\
%  & Exit 8 & 0.409 \scriptsize(\ensuremath{-}0.8) & 0.272 \scriptsize(\ensuremath{-}0.9) & 0.518 \scriptsize(\ensuremath{-}0.8) & 0.519 \scriptsize(\ensuremath{-}0.4) & 0.661 \scriptsize(\ensuremath{-}2.9) & 0.252 \scriptsize(0.0) \\
%  & Exit 12 & 0.542 \scriptsize(\ensuremath{-}0.5) & 0.336 \scriptsize(0.0) & 0.585 \scriptsize(\ensuremath{-}2.9) & 0.542 \scriptsize(\ensuremath{-}2.6) & 0.703 \scriptsize(\ensuremath{-}2.0) & 0.253 \scriptsize\textcolor{ForestGreen!60!black}{(+0.3)} \\
%  & \cellcolor{gray!12}Full (16) & \cellcolor{gray!12}0.605 \scriptsize(\ensuremath{-}0.7) & \cellcolor{gray!12}0.388 \scriptsize\textcolor{ForestGreen!60!black}{(+0.1)} & \cellcolor{gray!12}0.694 \scriptsize\textcolor{ForestGreen!60!black}{(+0.3)} & \cellcolor{gray!12}0.552 \scriptsize(\ensuremath{-}4.0) & \cellcolor{gray!12}0.737 \scriptsize(\ensuremath{-}1.0) & \cellcolor{gray!12}0.258 \scriptsize\textcolor{ForestGreen!60!black}{(+0.2)} \\
%  & No-Balcony & 0.606 \scriptsize(\ensuremath{-}0.6) & 0.394 \scriptsize\textcolor{ForestGreen!60!black}{(+0.7)} & 0.689 \scriptsize(\ensuremath{-}0.1) & 0.573 \scriptsize(\ensuremath{-}1.9) & 0.742 \scriptsize(\ensuremath{-}0.5) & 0.256 \scriptsize(0.0) \\
% \midrule
\multirow{5}{*}{7B}
 & Exit 15 & 0.612 & 0.324 & 0.482 & 0.632 & 0.718 & 0.330 \\
 & Exit 18 & 0.657 & 0.342 & 0.485 & 0.653 & 0.726 & 0.326 \\
 & Exit 21 & 0.685 & 0.371 & 0.489 & 0.654 & 0.740 & 0.335 \\
 & \cellcolor{gray!12}Full (32) & \cellcolor{gray!12}0.740 & \cellcolor{gray!12}0.407 & \cellcolor{gray!12}0.572 & \cellcolor{gray!12}0.696 & \cellcolor{gray!12}0.773 & \cellcolor{gray!12}0.384 \\
\midrule
\multirow{5}{*}{8B}
 & Exit 15 & 0.515 & 0.323 & 0.485 & 0.644 & 0.638 & 0.306 \\
 & Exit 18 & 0.585 & 0.361 & 0.560 & 0.667 & 0.668 & 0.589 \\
 & Exit 21 & 0.629 & 0.401 & 0.617 & 0.713 & 0.708 & 0.598 \\
 & \cellcolor{gray!12}Full (32) & \cellcolor{gray!12}0.785 & \cellcolor{gray!12}0.515 & \cellcolor{gray!12}0.784 & \cellcolor{gray!12}0.729 & \cellcolor{gray!12}0.799 & \cellcolor{gray!12}0.605 \\
\bottomrule
\end{tabular}}
\end{table}

\subsection{Relating Depth, Optimization, Utility, and Jailbreak Success}\label{subsec:synthesis}

Bringing together the results, we can now relate inference depth, attack optimizability, downstream utility, and jailbreak success within a single picture, and ask whether the robustness of shallow exits reflects a genuine safety property or is simply a byproduct of reduced capability.

\textbf{Depth couples capability, attackability, and vulnerability, but the strength of this coupling is backbone-dependent.} For Balcony-LLaMA3-8B, the three quantities move together tightly: clean utility, ease of GCG optimization (i.e., how low the attack loss falls), and ASR all increase monotonically and substantially from exit-15 to the full model. Exit-15 is simultaneously the least capable, the hardest to optimize against, and the least jailbreakable exit, while the full model is the most capable, the easiest to optimize against, and the most jailbreakable. This alignment suggests that, for LLaMA3, the apparent robustness of shallow exits is closely tied to how little task-relevant and instruction-following capability has been formed at that depth, rather than to a distinct, depth-localized safety mechanism.

For Balcony-LLaMA2-7B, this coupling is markedly weaker. Clean utility still increases monotonically with depth, but ASR is already high at exit-15 and increases only modestly toward the full model, and the GCG loss curves for all four exits converge to a narrow band within the first 50 iterations (Section~\ref{subsec:attack_optimization}). In other words, LLaMA2's shallow exits are attackable almost as easily as its full model, despite having visibly lower utility. This indicates that whatever representations GCG exploits to induce jailbreaks are already largely in place early in the network for this backbone, well before the model reaches its full downstream capability.

\textbf{Attack optimizability tracks jailbreak success more consistently than it explains it.} Across both models, the ordering of GCG loss across exits (full $<$ exit-21 $<$ exit-18 $<$ exit-15) mirrors the ordering of ASR, suggesting that harder-to-optimize exits are, unsurprisingly, harder to jailbreak. However, the \emph{magnitude} of the gap differs sharply between backbones: LLaMA3's exit-15 is both much harder to optimize against and much more robust in absolute terms than LLaMA2's exit-15, even though both models use the same exit configuration, training corpus, and self-distillation objective. This rules out training-side confounds as the source of the difference and points instead to how each backbone's pretraining distributes alignment-relevant computation across depth.

\textbf{Jailbreak success is largely decoupled from utility degradation.} Table~\ref{tab:utility_attacked} shows that, across nearly every exit and benchmark, adversarial suffixes optimized to induce unsafe generations leave clean-task utility essentially unchanged relative to Table~\ref{tab:utility_clean}, with ARC-E on the 7B model as the main exception. This indicates that GCG suffixes are acting in a targeted way, i.e., steering the model toward affirmative, unsafe completions, rather than degrading the model's general representations. Consequently, utility cannot be used as a proxy for robustness: an exit can be simultaneously low-capability and highly jailbreakable (LLaMA2-7B, exit-15) or, conversely, an exit's low ASR cannot be dismissed as merely reflecting a broken or degenerate sub-model, since its clean utility remains substantial.

% ============================================================
% TABLE 3: Main Attack Results
% ============================================================

% \begin{table}[t]
% \centering
% \caption{GCG attack results on both benchmarks. Final loss is the mean
% cross-entropy GCG minimizes over 50 behaviors after 150 steps. ASR is the
% share of those behaviors the HarmBench judge scores above 0.5. Each model is
% listed for every exit, the full model, and its no-balcony baseline.}
% \label{tab:attack_results}
% \small
% \renewcommand{\arraystretch}{1.15}
% \begin{tabular}{ll cc cc}
% \toprule
%  & & \multicolumn{2}{c}{\textbf{HarmBench}} & \multicolumn{2}{c}{\textbf{AdvBench}} \\
% \cmidrule(lr){3-4} \cmidrule(lr){5-6}
% \textbf{Model} & \textbf{Exit} & \textbf{Loss} & \textbf{ASR (\%)} & \textbf{Loss} & \textbf{ASR (\%)} \\
% \midrule
% \multirow{5}{*}{7B}
%  & Exit 15 & 0.479 & 36.0 & 0.445 & 46.0 \\
%  & Exit 18 & 0.455 & 40.0 & 0.490 & \textbf{54.0} \\
%  & Exit 21 & 0.420 & 40.0 & 0.426 & 48.0 \\
%  & \cellcolor{gray!12}Full (32) & \cellcolor{gray!12}0.389 & \cellcolor{gray!12}\textbf{44.0} & \cellcolor{gray!12}0.424 & \cellcolor{gray!12}42.0 \\
% \midrule
% \multirow{5}{*}{8B}
%  & Exit 15 & 1.904 & 0.0 & 1.530 & 4.0 \\
%  & Exit 18 & 1.203 & 4.0 & 1.021 & 2.0 \\
%  & Exit 21 & 1.057 & 10.0 & 0.906 & 8.0 \\
%  & \cellcolor{gray!12}Full (32) & \cellcolor{gray!12}0.501 & \cellcolor{gray!12}\textbf{24.0} & \cellcolor{gray!12}0.521 & \cellcolor{gray!12}\textbf{26.0} \\
% \bottomrule
% \end{tabular}
% \end{table}
\section{Conclusion}
In this paper, we explored the jailbreak vulnerability of depth-based dynamic inference in LLMs. We conducted this investigation through extensive experiments on two dynamic models (Balcony-LLaMA2-7B and Balcony-LLaMA3-8B) adapted with the Balcony method, using GCG as a representative jailbreak attack. Our findings are as follows: 1) shallow exits are consistently more robust to jailbreak attacks than deeper exits; 2) exit depth affects how easily an adversarial suffix can be optimized, with deeper exits reaching lower attack loss; and 3) adversarial suffixes affect utility uniformly across exit depths. These findings are associative and come from a single attack with fixed exit points; extending the study to routing policies, stronger attacks, and analyses of internal representations is left for future work.

\bibliographystyle{plain}
\bibliography{references}

% References follow the acknowledgments in the camera-ready paper. Use unnumbered first-level heading for
% the references. Any choice of citation style is acceptable as long as you are
% consistent. It is permissible to reduce the font size to \verb+small+ (9 point)
% when listing the references.
% Note that the Reference section does not count towards the page limit.
% \medskip

% {
% \small

% [1] Alexander, J.A.\ \& Mozer, M.C.\ (1995) Template-based algorithms for
% connectionist rule extraction. In G.\ Tesauro, D.S.\ Touretzky and T.K.\ Leen
% (eds.), {\it Advances in Neural Information Processing Systems 7},
% pp.\ 609--616. Cambridge, MA: MIT Press.

% [2] Bower, J.M.\ \& Beeman, D.\ (1995) {\it The Book of GENESIS: Exploring
%   Realistic Neural Models with the GEneral NEural SImulation System.}  New York:
% TELOS/Springer--Verlag.

% [3] Hasselmo, M.E., Schnell, E.\ \& Barkai, E.\ (1995) Dynamics of learning and
% recall at excitatory recurrent synapses and cholinergic modulation in rat
% hippocampal region CA3. {\it Journal of Neuroscience} {\bf 15}(7):5249-5262.
% }

%%%%%%%%%%%%%%%%%%%%%%%%%%%%%%%%%%%%%%%%%%%%%%%%%%%%%%%%%%%%

\appendix

% \section{Technical appendices and supplementary material}
% Technical appendices with additional results, figures, graphs, and proofs may be submitted with the paper submission before the full submission deadline (see above). You can upload a ZIP file for videos or code, but do not upload a separate PDF file for the appendix. There is no page limit for the technical appendices. 

% Note: Think of the appendix as ``optional reading'' for reviewers. The paper must be able to stand alone without the appendix; for example, adding critical experiments that support the main claims to an appendix is inappropriate. 

% %%%%%%%%%%%%%%%%%%%%%%%%%%%%%%%%%%%%%%%%%%%%%%%%%%%%%%%%%%%%

% \newpage
% \input{checklist.tex}

\end{document}